\documentclass[letterpaper, 10 pt, conference]{ieeeconf}  
\usepackage{graphicx}
\usepackage{multirow}
\usepackage{booktabs}
\usepackage{tabularx}
\usepackage{amsfonts}
\usepackage{cite}  
\usepackage{amsmath}
\usepackage{xcolor}
\usepackage{tcolorbox}
\usepackage{listings}
\usepackage{tabularx, booktabs, multirow, array}
\usepackage{marvosym}

\newcolumntype{Y}{>{\centering\arraybackslash}X}

\IEEEoverridecommandlockouts                              

\definecolor{blue}{rgb}{0,0,1}
\definecolor{sred}{rgb}{1,0,0}
\definecolor{green}{rgb}{0,.5,0}
\definecolor{orange}{rgb}{0.75, 0.4, 0}
\definecolor{teal}{rgb}{0.67, 0.80, 0.91}
\definecolor{purple}{rgb}{0.65,0,0.65}

\title{\LARGE \bf
IMR-LLM: Industrial Multi-Robot Task Planning and Program Generation using Large Language Models
}

\author{Xiangyu Su$^{1,2}$, Juzhan Xu$^{1,2}$, Oliver van Kaick$^{3}$, Kai Xu$^{4, \textsuperscript{\Letter}}$, Ruizhen Hu$^{1,\textsuperscript{\Letter}}$
\thanks{$^{1}$Shenzhen University. $^{2}$SpeedBot Robotics Co., Ltd. $^{3}$Carleton University. $^{4}$Institute of AI for Industries, Chinese Academy of Sciences.}%
\thanks{This work was done during an internship at SpeedBot Robotics Co., Ltd. \textsuperscript{\Letter} Kai Xu and Ruizhen Hu are corresponding authors.}%
}

\begin{document}

\maketitle
\thispagestyle{empty}
\pagestyle{empty}

\begin{abstract}
In modern industrial production, multiple robots often collaborate to complete complex manufacturing tasks. 
Large language models (LLMs), with their strong reasoning capabilities, have shown potential in coordinating robots for simple household and manipulation tasks. 
However, in industrial scenarios, stricter sequential constraints and more complex dependencies within tasks present new challenges for LLMs.
To address this, we propose IMR-LLM, a novel LLM-driven Industrial Multi-Robot task planning and program generation framework.
Specifically, we utilize LLMs to assist in constructing disjunctive graphs and employ deterministic solving methods to obtain a feasible and efficient high-level task plan. 
Based on this, we use a process tree to guide LLMs to generate executable low-level programs.
Additionally, we create IMR-Bench, a challenging benchmark that encompasses multi-robot industrial tasks across three levels of complexity.
Experimental results indicate that our method significantly surpasses existing methods across all evaluation metrics.
\end{abstract}

\section{INTRODUCTION}
\begin{figure}[!tbp]
        \centering
        \includegraphics[width=1.0\linewidth]{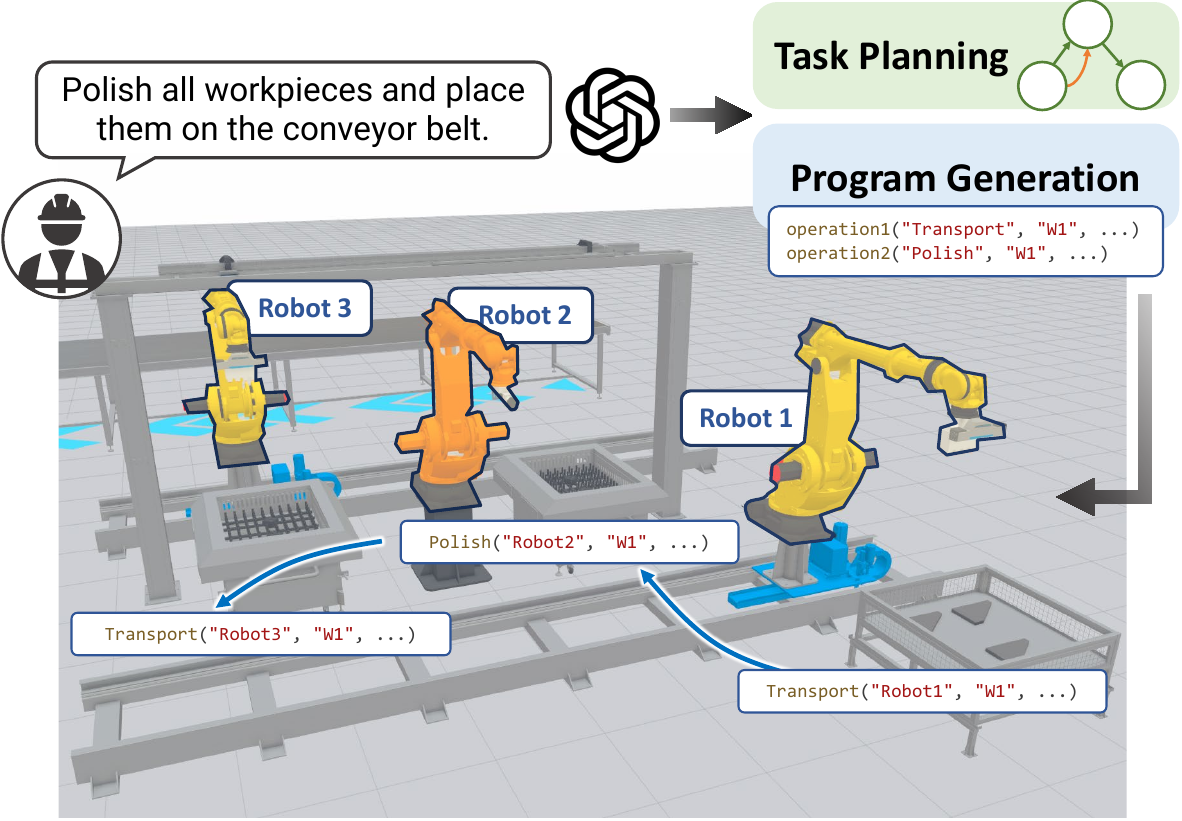}
        \vspace{-0.6cm}
	\caption{\textbf{A multi-robot industrial production line. }
    Our method transforms manufacturing tasks described in natural language into high-level task plans and low-level execution programs, allowing multiple robots to collaborate efficiently in completing the tasks.}
        \vspace{-0.6cm}
	\label{fig:teaser}
\end{figure}

In the rapidly evolving landscape of intelligent manufacturing, industrial embodied intelligence \cite{fan2025embodied, ren2024embodied, jqr-47-4-581} —robots equipped with autonomous perception, reasoning, and execution capabilities—is emerging as a cornerstone for next-generation manufacturing systems. 
Modern manufacturing tasks often require multiple robots to execute complex workflows collaboratively and effectively, which raises challenges for both high-level task planning and low-level program generation.
Recently, large language models (LLMs) have demonstrated remarkable potential in reasoning and code generation \cite{10.1145/3747588, 10403378}.
In this work, we focus on leveraging LLMs to solve the multi-robot task planning and program generation problem in industrial production lines, as illustrated in Fig. \ref{fig:teaser}.

Task planning typically comprises three sub-problems: decomposition, allocation, and scheduling, which respectively address breaking down tasks, assigning robots, and ordering execution of sub-tasks.
Existing multi-robot task planning approaches mainly focus on indoor household tasks \cite{kannan2024smart, obata2024lip, liu2024coherent, zhang2025lamma}, where LLMs are used directly to determine the execution order of sub-tasks.
However, in industrial production lines, tasks are often decomposed into standardized operations to meet specific production requirements. 
These operations are characterized by more rigid execution sequences and intricate dependencies. 
As a result, relying solely on the reasoning capabilities of LLMs may not be sufficient to ensure feasible or optimal planning.
Program generation aims to produce execution code that enables robots to complete allocated tasks.
Existing methods typically rely on few-shot prompting, where a handful of task-specific examples are directly provided to LLMs to guide code generation \cite{kannan2024smart, singh2022progprompt}. 
The generated programs may overfit the specific examples and fail to align with the actual execution environment, leading to low executability \cite{raman2022planning}.
In our setting, while the types of operations are predefined and the execution procedures within the same type of operations remain relatively consistent, variations in implementation still persist across different production lines. 
Consequently, program generation must not only encompass all critical process steps without omissions but also adapt to production-specific constraints to ensure the generated code is executable in practice.

To address these challenges, we propose IMR-LLM, an LLM-driven framework for multi-robot task planning and program generation in industrial production lines. 
During the task planning phase, given the motivation that the sequence of operations and resource conflict constraints during task execution can be effectively modeled using disjunctive graphs \cite{10.1287/opre.17.6.941} and solved with deterministic algorithms \cite{liu2006new, sobeyko2016heuristic}, we utilize LLMs to assist in constructing disjunctive graphs from a natural language task description, and integrate these with existing solving methods to generate accurate and efficient scheduling plans. 
During the program generation phase, we observed that tree structures more effectively represent the execution sequences within operations and account for implementation variations caused by different environments, thus providing clearer guidance for the generation process. 
Accordingly, we employ LLMs to derive an operation process tree that encompasses all types of operations from given examples, replacing the reliance on specific examples in existing methods and thereby enhancing the executability of the generated programs.

The contributions of this paper are as follows:
\begin{itemize}
    \item We introduce \textbf{IMR-LLM}, a multi-robot task planning and program generation framework in industrial production lines that integrates LLMs with heuristic algorithms to construct and solve disjunctive graphs, while leveraging a process tree to guide program generation.
    \item We create \textbf{IMR-Bench}, a benchmark designed to evaluate the performance of multi-robot systems in industrial tasks, which includes manufacturing tasks of varying complexity and meticulously designed metrics.
    \item We implement and evaluate our framework in both simulated and real-world settings, performing thorough testing across a wide array of tasks.
\end{itemize}

\section{RELATED WORK}

\subsection{Multi-Robot Task Planning}
Task planning addresses the challenge of determining ``when each robot should complete which part of the task'' in multi-robot collaboration to ensure overall efficient execution. 
Existing LLM-based methods can be categorized into decentralized and centralized approaches. In decentralized methods \cite{mandi2024roco, zhang2023building, liu2024leveraging, 10.1145/3676641.3716016}, each robot is equipped with an LLM agent that determines actions by exchanging capabilities and observations through dialogue. 
However, as the number of robots and dialogue rounds increase, the prompt length rapidly expands, leading to unstable performance due to insufficient long-context reasoning \cite{10610676}. 
In contrast, centralized methods, where a single LLM handles planning, have been proven superior in terms of success rate and token efficiency \cite{10610676}. 
Existing methods \cite{kannan2024smart, zhang2025lamma, obata2024lip, liu2024coherent, khan2025safety, wan2025embodiedagent} typically decompose the task planning problem into three sub-problems: decomposition, allocation, and scheduling. 
COHERENT \cite{liu2024coherent} addresses all sub-problems in a single call, while SMART-LLM \cite{kannan2024smart} first decomposes the task and determines the execution order, then allocates robots to subtasks. LaMMA-P \cite{zhang2025lamma} first decomposes and allocates robots, then determines the execution order, differing only in the timing of scheduling resolution, but both rely on LLMs to directly generate the order. 
LiP-LLM \cite{obata2024lip} further improves by not directly outputting the execution order but by constructing a dependency graph through analyzing subtask dependencies, which is used for robot allocation and scheduling, ensuring execution of specific subtasks occurs only after all preceding dependencies have been completed.

Currently, these methods are mostly applied in indoor scenes for household or manipulation tasks, where the execution order is relatively flexible. 
For example, when making a sandwich, you can cut vegetables first and then fry the patty, or do both simultaneously. 
However, in industrial scenes, there are stricter execution sequences between operations, and due to mutual-exclusion constraints on robot and machine access, the dependencies between operations are more complex. Therefore, relying solely on the capabilities of large models may not yield a feasible and efficient schedule.

\subsection{Robot Program Generation}
Program generation addresses the question of ``how each robot should complete the allocated task.'' 
This process typically involves providing a scene description, task description, and an atomic skill library, leveraging the code generation capabilities of large models to produce executable code for robots to perform and complete tasks in the target environment. 
Existing methods \cite{huang2022language, raman2022planning, codeaspolicies2022, singh2022progprompt} commonly offer the large model specific instruction-to-code pairs to guide generation using examples. ProgPrompt \cite{singh2022progprompt} introduced a prompting scheme that enables LLMs to generate Python code composed of robotic arm action primitives while incorporating environmental state feedback. This prompting method is also employed by SMART-LLM \cite{kannan2024smart} to generate underlying execution code. Code-as-Policies \cite{codeaspolicies2022} generates Python code containing perception and control APIs, which is directly utilized as the policy for robots to complete tasks. Although these approaches may provide suitable code for indoor environments, they may fail to align with the actual execution environment in constrained industrial scenes.

\section{DEFINITIONS}
\begin{figure*}[t]
  \centering
  \includegraphics[width=1\textwidth]{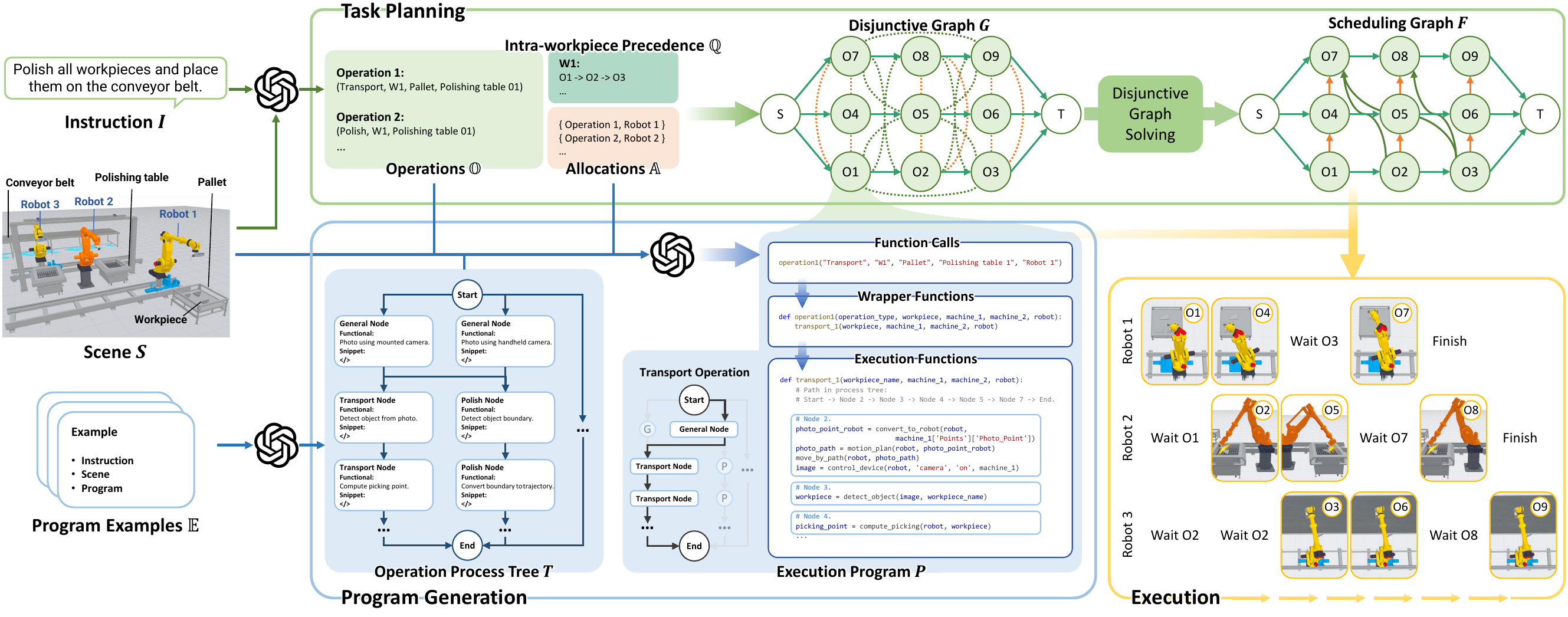}
  \vspace{-0.6cm}
  \caption{\textbf{An overview of our method.} Given an instruction $I$, an industrial scene $S$, and program examples $\mathbb{E}$, our method performs task planning (highlighted in green) to decompose operations, assign robots, and schedule operations using a disjunctive graph and a heuristic solver. This is followed by program generation (highlighted in blue) that translates the plan into executable Python code under the guidance of an operation process tree. The resulting high-level plan and low-level program enable collaborative execution by multiple robots.
  }
  \vspace{-0.6cm}
  \label{fig:overview}
\end{figure*}
\subsection{Disjunctive Graph in Robotic Job-Shop Scheduling}
Job-Shop Scheduling Problem (JSSP) \cite{sellers1996survey, xiong2022survey} is a classic and highly challenging combinatorial optimization problem frequently encountered in manufacturing and automated production.
The primary objective is to determine the optimal processing sequence for a set of jobs, where each job consists of multiple operations performed on different machines. 
In this paper, we focus on the Robotic Job-Shop Scheduling Problem (RJSSP) \cite{sun2021novel, wen2023green}. 
Unlike traditional JSSP, RJSSP introduces an additional complexity by requiring not only mutually exclusive access to manufacturing machines (such as polishing tables) but also the exclusive allocation of robots, which serve as a limited and shared resource.

The disjunctive graph \cite{10.1287/opre.17.6.941} is widely utilized in JSSP to explicitly represent both operation sequencing constraints and resource conflict constraints, serving as a foundational data structure for accurately defining problem inputs and solutions.
For RJSSP, we adopt an extended disjunctive graph, where the graph $G = \{\mathbb{V}, \mathbb{C}, \mathbb{D_M}, \mathbb{D_R}\}$ consists of the following elements:

\begin{itemize}
    \item Vertex set $\mathbb{V}$.  Each vertex $V$ represents an operation $O$.
    $O$ is further defined by its operation type $O_T$, workpiece $W$, associated machines $M_1$ and $M_2$, where $M_2$ is applicable only for transport operations.
    Note that we focus on five common types of operations typically seen in industrial production: transport, polishing, welding, beveling, and assembly. 
    Additionally, $\mathbb{V}$ includes a source vertex $V_S$ and a terminal vertex $V_T$, representing the start and end of all operations.
    \item Conjunctive arcs $\mathbb{C}$. Each arc $C$ is a directed edge that represents precedence constraints among operations within the same workpiece.
    \item Machine disjunctive arcs $\mathbb{D_M}$. Each arc $D_M$ is an undirected edge that represents conflicts arising when different operations compete for the same manufacturing machine that cannot be used simultaneously.
    \item Robot disjunctive arcs $\mathbb{D_R}$. Each arc $D_R$ is an undirected edge that represents conflicts arising when different operations compete for the same robot.
\end{itemize}
Given a disjunctive graph, the core problem is to determine the direction for all undirected edges and ensure that the graph remains acyclic, which represents a solution to the RJSSP.
Among all possible orientation schemes, the goal is to find the optimal solution that minimizes an objective function (such as minimizing completion time) through search or optimization. 
In this paper, we call the process of this optimization ``solving the disjunctive graph'', for short.
Common solution methods include constraint programming \cite{green2024using}, heuristic algorithms \cite{liu2006new, sobeyko2016heuristic,wang2024multi}, and machine learning-based approaches \cite{shao2022adaptive, liu2025flexible}.

\subsection{Problem Formulation}
Given an industrial production scene $S$, a user instruction $I$, and program examples $\mathbb{E}$, our goal is to generate the high-level task plan $H$ and low-level execution program $P$ needed to complete the instruction. 
The plan $H$ specifies all operations required to fulfill the instruction, the robots involved in execution, and the sequence in which they will be carried out, optimizing the utilization of robots and machines through parallel execution wherever possible. 
The program $P$ defines the execution method for each operation. 
With both $H$ and $P$, we can drive the robots to complete the given instruction in a simulated or real environment.

The input scene is defined as \( S = \{\mathbb{R}, \mathbb{M}, \mathbb{W}\} \), where \( \mathbb{R} = \{ R^i \}_{i=1}^X \) represents the set of robots, each associated with controllable external devices (e.g., magnetic gripper, welding gun) and the machines within their workspace; \( \mathbb{M} = \{ M^i \}_{i=1}^Y \) represents the set of machines, characterized by their name (e.g., polishing table), the workpieces placed on them, and their accessibility by multiple robots; and \( \mathbb{W} = \{ W^i \}_{i=1}^Z \) represents the set of workpieces, described by their type and the sequence of states they undergo during processing (e.g., polished, welded). Program examples are defined as \( \mathbb{E} = \{\{I_E^j, S_E^j, P_E^j\}_{j=1}^K \}\), where $I_E^j$, $S_E^j$, and $P_E^j$ denote example instruction, scene, and program, respectively.

The output high-level plan is defined as $H=\{\mathbb{O},\mathbb{A},F\}$.
Here, \( \mathbb{O} = \{ O^i \}_{i=1}^L \) represents the set of all operations necessary to complete the instruction.
$\mathbb{A}=\{\{O^i, R^j\}_{i=1}^{L} \mid R^j \in \mathbb{R}\}$ is the set of allocations, where $\{O^i, R^j\}$ indicates that operation $O^i$ is allocated to robot $R^j$.
In addition, we use $\mathbb{Q} = \left\{ \{ W^i, [O^j, O^k, \dots, O^l] \} \mid W^i \in \mathbb{W}, O \in \mathbb{O} \right\}$ to indicate the order of processing operations for the same workpiece. 
Based on $\mathbb{A}$, $\mathbb{O}$ and $\mathbb{Q}$, we construct a disjunctive graph $G$. 
Solving the disjunctive graph yields a feasible schedule graph $F$, which illustrates the execution sequence of all operations.
A program is defined as $P=\{\mathbb{P_C}, \mathbb{P_W},\mathbb{P_E}\}$, where \( \mathbb{P_C} = \{ P_C^i \}_{i=1}^L \) represents the set of function calls, serving as the entry point for executing operations; \( \mathbb{P_W} = \{ P_W^i \}_{i=1}^L \) represents the set of wrapper functions, each of which indicates the execution function that can be used to execute the operations; \( \mathbb{P_E} = \{ P_E^i \}_{i=1}^B \) represents the set of execution functions. 
Note that $B$ is often less than $L$ because some operations may share the same execution function but have different call parameters.  
Each executable statement in $P_E$ represents an atomic skill from a skill library, such as moving to a specified position, recognizing target objects in an image, or controlling the start and stop of external devices, where all skills have been pre-implemented.

\section{METHOD}
Our framework, IMR-LLM, is specifically designed to tackle the challenges of multi-robot task planning and program generation in industrial settings.
Fig. \ref{fig:overview} offers an overview of our framework, which is composed of three modules: task planning, program generation, and execution.

\subsection{Task Planning}
The initial step of our method involves analyzing the input scene and instruction, decomposing the instructions into operations, allocating robots, and determining schedules. A disjunctive graph, a specialized data structure for job shop scheduling problems, is constructed using the reasoning capabilities of LLMs. After that, we employ well-established methods to solve the graph, ensuring accurate and efficient task planning.

Given the scene $S$ and instruction $I$ as inputs, we leverage the LLM's strong reasoning capabilities via Chain of Thought (CoT) \cite{wei2022chain} to infer implicit constraints and simultaneously generate a complete operation set $\mathbb{O}$, allocation set $\mathbb{A}$, and intra-workpiece precedence set $\mathbb{Q}$. 
Note that unlike previous methods \cite{kannan2024smart, zhang2025lamma} that depend on few-shot examples to prompt the model in plan generation, our approach employs a set of general rules. 
Our experiments indicate that this rule-based prompting method achieves superior generalization across various tasks and scenes.
Subsequently, the textual outputs are converted into a disjunctive graph $G$. 
Each operation in $\mathbb{O}$ is represented as a vertex $V$ in $G$. 
Intra-workpiece precedences $\mathbb{Q}$ are transformed into conjunctive arcs $\mathbb{C}$. 
The undirected disjunctive edges ${\mathbb{D_M}}$ and $\mathbb{D_R}$ are set when two operations require the same machine or robot with mutually exclusive access.

After constructing $G$, existing methods \cite{liu2006new, sobeyko2016heuristic} can be employed to solve the scheduling problem to get the scheduling graph $F$, which represents a feasible execution order of all operations.
In our work, we use a heuristic First-In-First-Out (FIFO) \cite{nasri2018fifo} algorithm to solve the disjunctive graph. 
If multiple operations share mutually exclusive resources, the execution order is determined by their sequence in the operation set $\mathbb{O}$.
Note that other solving methods can also be used to meet different objective requirements.

Compared to previous methods \cite{zhang2025lamma, liu2024coherent, kannan2024smart, obata2024lip} that rely entirely on LLMs for solving scheduling problems, our approach only requires the LLM to determine the sequence of operations for individual workpieces. 
This restriction means the LLM does not need to handle complex resource conflicts, thereby simplifying the analysis and generation processes. 
Additionally, using disjunctive graph parsing to solve the scheduling problem enhances explainability, which assists engineers in reviewing and verifying outcomes in industrial workshop scenarios where high efficiency and accuracy are essential.

\subsection{Program Generation}
Upon obtaining the high-level task plan $H$, we employ an LLM to generate a low-level program $P$.
Instead of using existing methods which directly employ a few in-context learning program examples \cite{kannan2024smart} to prompt the LLM for generation, we have observed that identical types of operations usually follow a consistent and strictly sequential process. 
For example, polishing operations typically involve: photographing, detecting object boundary, computing trajectory, moving to the start point, polishing along the trajectory, and returning to the initial position. 
Despite this consistency, minor variations may occur due to different production line configurations—for instance, photographing differs when the camera is mounted on a bracket above the robot compared to when it is held by the robot.
Furthermore, some processes are shared across different operations, such as photographing for workpiece localization in transportation, which resembles the initial process of polishing.

Based on these observations, we find that a tree-structured representation offers significant advantages for organizing and encoding execution procedures. 
This representation clearly captures the fixed sequential dependencies within each operation and models diverse execution variants through branching structures. 
It also enables different operations to reuse general action steps, thereby reducing overall structural complexity. 
Consequently, we construct a unified process tree $T$ that encompasses all operations, serving as a replacement for in-context learning examples.
\begin{figure*}[!tbp]
        \centering
        \includegraphics[width=1.0\linewidth]{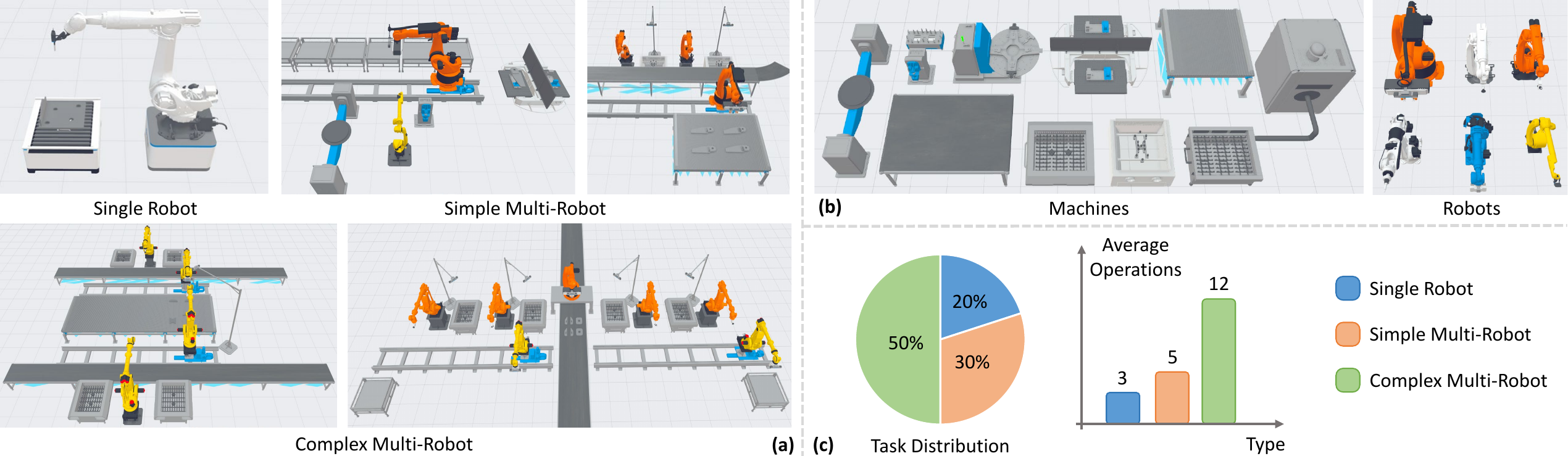}
        \vspace{-0.6cm}
	\caption{\textbf{An overview of our dataset.} Our tasks consist of (a) various scenes and (b) various machines and robots equipped with different end-effectors. (c) A pie chart showing the distribution of task types on the left and a bar chart showing the average number of operations per task type on the right.}
        \vspace{-0.6cm}
	\label{fig:dataset}
\end{figure*}

The operation process tree $T$ is synthesized from the program examples $\mathbb{E}$ by an LLM. 
These examples encompass multiple instructions, scenes, and corresponding programs, covering all types of operations. 
To construct this tree, the LLM is prompted to comprehensively analyze all examples, segment the program according to functionality, where each segment corresponds to a node in the tree. 
Then, it analyzes all nodes with identical functionality; if they have the same implementation, they are merged into a single node. 
Otherwise, it analyzes the corresponding scene to further refine the functional description.
After the generation is complete, we manually verify the tree to ensure its correctness. 
Note that since the tree includes the complete processes for all types of operations, we construct and validate it only once and employ it in all testing scenes and tasks.
Each node in $T$ has a unique index, type, functional description, and snippet. 
The index identifies the node, the type specifies supported operations (``general'' if multiple), the description outlines its function and conditions, and the snippet gives the steps to realize it. An example of a node representation is:
\begin{tcolorbox}[colback=gray!5!white, colframe=teal, width=0.485\textwidth, arc=3mm,auto outer arc, sharp corners, left=1mm, top=1mm, bottom=1mm, right=1mm]

{\renewcommand{\baselinestretch}{0.8} \scriptsize
\vspace{-2mm}
\lstset{
    basicstyle=\ttfamily\scriptsize,  
    keywordstyle=\bfseries,           
    morekeywords={Index, Type, Functional, description, Snippet}  
}
\begin{lstlisting}[basicstyle=\ttfamily, keywordstyle=\bfseries]
Index: 2
Type: General
Functional description: 
Photo using handheld camera.
Snippet:
photo_point_robot = convert_to_robot(robot,
                     machine_1['Points']['Photo_Point'])
photo_path = motion_plan(robot, photo_point_robot)
move_by_path(robot, photo_path)
image = control_device(robot, 'camera', 'on', machine_1)
\end{lstlisting}
\vspace{-2mm}}
\end{tcolorbox}
\vspace{-1mm}
Once the process tree is obtained, we use it to guide program generation in a Python code generation context \cite{singh2022progprompt}.
Specifically, given the operation process tree $T$ in JSON format, scene $S$, operation set $\mathbb{O}$ and allocation set $\mathbb{A}$ as input, the LLM outputs a complete program $P$. 
During the generation process, the LLM is first tasked with inferring a unique branch from start to finish in the tree for each operation based on the scene. 
Then, it combines the snippets contained within all nodes of the branch to obtain a general execution function for each branch. 
Finally, for each operation, it creates a wrapper function that calls the general execution function. 
By reusing general execution functions for operations with the same process, this approach reduces code redundancy and eases the LLM's generation workload.

The use of process tree transforms the code generation problem into a path selection problem.
Since execution differences caused by various scenes are clearly represented by distinct nodes in the tree, ambiguous execution logic is also avoided, thereby enhancing the completeness and executability of the code. 
Moreover, the tree structure is highly scalable, allowing for the support of new operation types by simply adding or modifying nodes, thereby achieving modularity and reusability.

\subsection{Execution}
After obtaining the task plan $H$ and execution code $P$, we can drive the robot in the scene to complete the specified task. The feasible scheduling graph $F$ in $H$ determines the execution sequence of operations; an operation can only be initiated when all its preceding dependencies are fulfilled. 
Once an operation is initiated, its execution is transferred to the corresponding execution function via function calls in $P$.


\section{EXPERIMENTS}
\begin{table*}[!tbp]
\centering
\footnotesize
\caption{Evaluation of IMR-LLM and baselines on different categories of tasks in the IMR-Bench dataset.}
\vspace{-0.28cm}
\begin{tabularx}{\textwidth}{l*{15}{Y}}
\toprule
\multirow{2}{*}{\textbf{Methods}} &
\multicolumn{5}{c}{\textbf{Single Robot}} &
\multicolumn{5}{c}{\textbf{Simple Multi-Robot}} &
\multicolumn{5}{c}{\textbf{Complex Multi-Robot}} \\
\cmidrule(lr){2-6} \cmidrule(lr){7-11} \cmidrule(lr){12-16}
& \textbf{OC~$\uparrow$} & \textbf{SE~$\uparrow$} & \textbf{Exe~$\uparrow$} & \textbf{GCR~$\uparrow$} & \textbf{SR~$\uparrow$}
& \textbf{OC} & \textbf{SE~$\uparrow$} & \textbf{Exe~$\uparrow$} & \textbf{GCR~$\uparrow$} & \textbf{SR~$\uparrow$}
& \textbf{OC~$\uparrow$} & \textbf{SE~$\uparrow$} & \textbf{Exe~$\uparrow$} & \textbf{GCR~$\uparrow$} & \textbf{SR~$\uparrow$} \\
\midrule
SMART-LLM \cite{kannan2024smart} &
0.83 & 0.70 & 0.50 & 0.50 & 0.50 &
0.67 & 0.46 & 0.46 & 0.32 & 0.20 &
0.50 & 0.04 & 0.00 & 0.03 & 0.00 \\
LaMMA-S \cite{zhang2025lamma} &
0.80 & 0.80 & 0.70 & 0.70 & 0.70 &
0.71 & 0.67 & 0.40 & 0.45 & 0.33 &
0.56 & 0.26 & 0.20 & 0.33 & 0.16  \\
LaMMA-O \cite{zhang2025lamma} &
0.80 & 0.80 & 0.80 & 0.80 & 0.80 &
0.71 & 0.67 & 0.53 & 0.55 & 0.46 &
0.56 & 0.26 & 0.28 & 0.37 & 0.20 \\
LiP-O  \cite{obata2024lip} &
\textbf{1.00} & \textbf{1.00} & 0.90 & 0.90 & 0.90 &
0.93 & 0.80 & 0.73 & 0.74 & 0.73 &
0.63 & 0.28 & 0.36 & 0.42 & 0.24 \\
\midrule
Ours (GPT-4o) &
\textbf{1.00} & \textbf{1.00} & 0.90 & 0.98 & 0.90 &
\textbf{1.00} & \textbf{1.00} & \textbf{0.87} & \textbf{0.94} & \textbf{0.87} &
\textbf{0.88} & \textbf{0.75} & \textbf{0.76} & \textbf{0.79} & \textbf{0.68} \\
Ours (Qwen3-32B)  &
\textbf{1.00} & \textbf{1.00} & \textbf{1.00} & \textbf{1.00} & \textbf{1.00} &
\textbf{1.00} & 0.93 & \textbf{0.87} & 0.90 & \textbf{0.87} &
0.85 & 0.71 & \textbf{0.76} & \textbf{0.79} & 0.60 \\

\bottomrule
\end{tabularx}
\vspace{-0.2cm}
\label{table:comparison}
\end{table*}

\begin{table*}[!tbp]
\centering
\footnotesize
\caption{Ablation study of different variations of IMR-LLM.}
\vspace{-0.28cm}
\begin{tabularx}{\textwidth}{l*{15}{Y}}
\toprule
\multirow{2}{*}{\textbf{Methods}} &
\multicolumn{5}{c}{\textbf{Single Robot}} &
\multicolumn{5}{c}{\textbf{Simple Multi-Robot}} &
\multicolumn{5}{c}{\textbf{Complex Multi-Robot}} \\
\cmidrule(lr){2-6} \cmidrule(lr){7-11} \cmidrule(lr){12-16}
& \textbf{OC~$\uparrow$} & \textbf{SE~$\uparrow$} & \textbf{Exe~$\uparrow$} & \textbf{GCR~$\uparrow$} & \textbf{SR~$\uparrow$}
& \textbf{OC~$\uparrow$} & \textbf{SE~$\uparrow$} & \textbf{Exe~$\uparrow$} & \textbf{GCR~$\uparrow$} & \textbf{SR~$\uparrow$}
& \textbf{OC~$\uparrow$} & \textbf{SE~$\uparrow$} & \textbf{Exe~$\uparrow$} & \textbf{GCR~$\uparrow$} & \textbf{SR~$\uparrow$} \\
\midrule
Ours ($w/ order$)
& \textbf{1.00} & \textbf{1.00} & \textbf{0.90} & 0.95 & \textbf{0.90}
& \textbf{1.00} & 0.67 & 0.47 & 0.42 & 0.47 
& 0.85 & 0.07 & 0.08  & 0.10 & 0.00 \\

Ours ($w/ dependency$)
& \textbf{1.00} & \textbf{1.00} & \textbf{0.90} & \textbf{0.98} & \textbf{0.90} 
& \textbf{1.00} & 0.73 & 0.80 & 0.80 & 0.60 
& 0.83 & 0.39 & 0.44  & 0.47 & 0.36 \\

Ours ($w/o \ T$)
& \textbf{1.00} & \textbf{1.00} & 0.80 & 0.80 & 0.80 
& \textbf{1.00} & \textbf{1.00} & 0.64 & 0.74 & 0.64 
& \textbf{0.88} & \textbf{0.75} & 0.48 & 0.53 & 0.44 \\

Ours 
& \textbf{1.00} & \textbf{1.00} & \textbf{0.90} & \textbf{0.98} & \textbf{0.90} 
& \textbf{1.00} & \textbf{1.00} & \textbf{0.87} & \textbf{0.94} & \textbf{0.87} 
& \textbf{0.88} & \textbf{0.75} & \textbf{0.76} & \textbf{0.79} & \textbf{0.68} \\
\bottomrule
\end{tabularx}
\vspace{-0.4cm}
\label{table:ablation}
\end{table*}
\subsection{Benchmark}
\textbf{Dataset. }As shown in Fig. \ref{fig:dataset}, we create a benchmark dataset, IMR-Bench. Built upon the KunWu platform \cite{speedbot25}, our benchmark comprises 23 scenes collected and adapted from real industrial environments by production line design experts. Each scene involves between 1 to 7 robots.
Based on these scenes, we developed 50 manufacturing tasks that reflect actual industrial needs. 
These tasks are categorized into three levels of difficulty: \textbf{single-robot tasks}, which involve only 1 robot and consist of up to 5 operations; \textbf{simple multi-robot tasks}, which involve up to 3 robots and encompass up to 10 operations, executed either in parallel or in sequence; and \textbf{complex multi-robot tasks}, which involve up to 7 robots and incorporate up to 24 operations, executed with a mix of parallel and sequential order.

In the dataset, the scenes $S$ are described using JSON format, which represent the robots, machines, and workpieces within each scene. These descriptions serve as input to our method along with the task instruction $I$ provided in text format. To evaluate the quality of the output, we manually create task-specific ground truth, which includes the operation decomposition $\mathbb{O}_{GT}$, allocation $\mathbb{A}_{GT}$, scheduling $S_{GT}$, and corresponding programs $P_{GT}$.

\textbf{Metrics.} We use five metrics to evaluate the quality of the generated results. For operation decomposition and allocation, we use \textit{Operation consistency (OC)}, which is calculated as the intersection-over-union between the generated $\mathbb{A}_{gen}$ and the GT allocations $\mathbb{A}_{GT}$. 
For scheduling, we use \textit{Scheduling Efficiency (SE)}, which is calculated as: 

\begin{equation}
\small
\textit{SE} =
\begin{cases}
0, &  \text{if } F \text{ is not feasible,} \\[5pt]
1, & \text{if } TS(F_{\text{GT}}) = TS(F_{\text{gen}}), \\[5pt]
\dfrac{|\mathbb{O}_{\text{GT}}| - TS(F_{\text{gen}})}{|\mathbb{O}_{\text{GT}}| - TS(F_{\text{GT}})}, 
& \text{otherwise},
\end{cases}
\end{equation}
where $TS(F)$ indicates the makespan of scheduling graph $F$. 
Note that \textit{SE} is calculated only when $OC=1$; otherwise, it is set to $0$.
For program generation, given the various reasonable ways to name variables in the program, we do not directly compare the generated $P_{gen}$ with $P_{GT}$. 
Instead, we use \textit{Executability (Exe)} \cite{zhang2025lamma} to determine whether the program can be executed, and \textit{Goal Condition Recall (GCR)} \cite{zhang2025lamma} to assess the extent to which the generated $P_{gen}$ accomplishes the given task. 
$GCR$ is calculated as:
\begin{equation}
\small
\textit{GCR} =
\frac{\;\;\text{Status}(P_{\text{gen}}) \cap \text{Status}(P_{\text{GT}})\;\;}
{\text{Status}(P_{\text{GT}})} ,
\end{equation}
where $Status$ represents the set of workpiece states in the scene that have changed relative to the initial state after executing $P$. 
These symbolic states indicate the positions of the workpieces and the operations they have undergone, such as being polished or welded.
In addition, we use the \textit{Success Rate (SR)} \cite{zhang2025lamma} to indicate whether the generated results have completed all operations under the optimal scheduling. When both \textit{SE} and \textit{GCR} are 1, \textit{SR} is 1; otherwise, it is 0.
\subsection{Simulation Experiments}

\textbf{Baselines.}
In the context of industrial settings, there are currently no readily available methods to address the problems of task planning and program generation. 
However, relevant methods exist for indoor scenes that can fully or partially accomplish our task. 
We selected three of the most pertinent methods and their variants for comparison:

\begin{itemize}
    \item \textbf{SMART-LLM} \cite{kannan2024smart} addresses both sub-problems simultaneously, using a different task planning sequence. 
    It determines the operation schedule during decomposition, followed by allocation, with task planning and program generation guided by specific examples.
    \item \textbf{LaMMA-P} \cite{zhang2025lamma} addresses only task planning in our setting, following the same sequence as our method but also using LLM to directly determine the operation schedule based on specific examples. Since its built-in symbolic solver cannot generate executable programs in our setting, we combined the program generation module from SMART-LLM\cite{kannan2024smart} and our method to create two variants, LaMMA-S and LaMMA-O.
    \item \textbf{LiP-LLM} \cite{obata2024lip} addresses only task planning, using a sequence that differs from ours. 
    It starts with operation decomposition, followed by schedule determination and linear programming for allocation. 
    Unlike previous methods that directly output the order of operations\cite{kannan2024smart, zhang2025lamma}, LiP-LLM uses LLM to analyze dependencies and construct a dependency graph. 
    The decomposition and dependency generation are guided by general rules as ours. 
    To obtain programs, we integrated our program generation module, resulting in a variant called LiP-O.

\end{itemize}

\textbf{Experimental Results.}
Tab. \ref{table:comparison} presents the quantitative results compared with the baselines on all 50 tasks in IMR-Bench. All baselines employ \texttt{GPT-4o} \cite{achiam2023gpt} to ensure a fair comparison.
SMART-LLM relies on in-context examples, exhibiting poor generalization that leads to redundant or missing operations and lower \textit{OC}. Moreover, it determines scheduling without allocation context, frequently causing conflicting assignments and infeasible plans (lower \textit{SE}), ultimately achieving the lowest \textit{SR}.

For LaMMA-S and LaMMA-O, scheduling is determined post-allocation, allowing the LLM to account for inter-robot constraints in schedule generation.
This results in an improvement in \textit{SE} compared to SMART-LLM. 
However, the method is still constrained by the direct generation of execution sequences through the LLM. As task complexity increases, the LLM struggles to produce reasonable execution sequences directly.
When comparing LaMMA-S and LaMMA-O, the use of operation process tree enhances \textit{Exe} and \textit{GCR}, although this improvement is limited due to most unexecuted cases arising from unreasonable planning.

LiP-O demonstrates significant performance improvements in single-robot and simple multi-robot tasks compared to the previous baselines and performs comparably to our method for single-robot tasks. 
In these two simple task types, allowing the LLM to first generate an operation dependency graph and then determining robot allocation based on this graph facilitates reasonable high-level planning. 
However, in complex multi-robot tasks, the more intricate execution sequence constraints and resource dependencies among operations result in the dependencies generated by the LLM often being incomplete. 
This incompleteness leads to allocation process failures and results in lower \textit{OC} and \textit{SE}, consequently causing a relatively low \textit{SR}.

\begin{figure*}[t]
  \centering
  \includegraphics[width=1\textwidth]{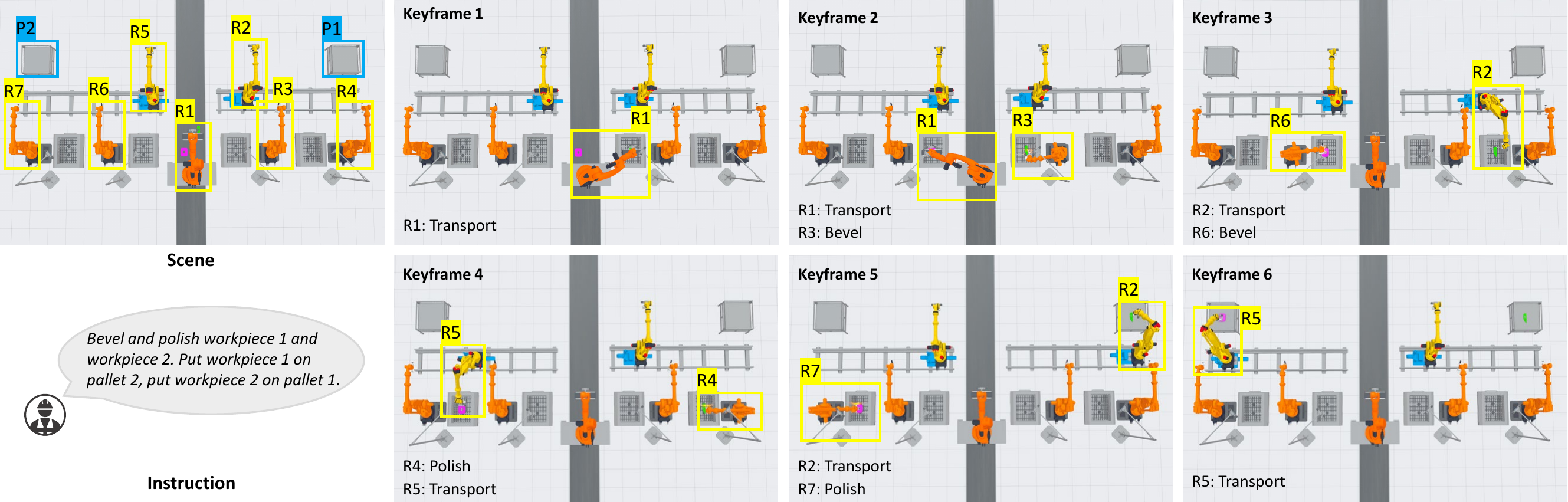}
  \vspace{-0.6cm}
  \caption{\textbf{Keyframes in the execution process.} Workpiece 1 and 2 (highlighted in pink and green) are initially placed on the conveyor belt. The robots (highlighted in the yellow boxes) collaborate to complete the operations and finally place workpiece 2 on pallet 1 (keyframe 5) and workpiece 1 on pallet 2 (keyframe 6). Each keyframe contains textual annotations describing the operations performed by the robots.
  }
  \vspace{-0.6cm}
  \label{fig:res}
\end{figure*}

In our method, the LLM only needs to analyze the sequence of operations within the workpiece. 
Complex dependencies are modeled and solved using a disjunctive graph, achieving the highest \textit{OC} and \textit{SE}. 
The use of the process tree further enhances the generated program's \textit{Exe} and \textit{GCR}, resulting in the highest \textit{SR}. 
Additionally, we employed the open-source \texttt{Qwen3-32B-thinking} \cite{yang2025qwen3} as the LLM to demonstrate the generalization capability of our method across models of different capacities; both models exhibited similar performance.
Fig. \ref{fig:res} illustrates an example where our method successfully and efficiently accomplishes the given task in 6 steps.

Despite the overall superior performance, we observed a decline in the \textit{SR} for complex multi-robot tasks. 
This is primarily because as task difficulty and scene complexity escalate, the LLM becomes prone to generating suboptimal or incorrect decompositions and allocations (e.g., assigning an operation to an out-of-reach robot), which directly leads to a decrease in \textit{OC}. 
Furthermore, the increased task complexity extends the context length required for program generation, increasing the likelihood of hallucinations or logical inconsistencies, resulting in lower \textit{Exe}. 
The combination of these factors ultimately contributes to the reduced overall \textit{SR}.

\textbf{Ablation Study.}
We evaluated different variants of our method to illustrate the impact of utilizing disjunctive graphs in task planning and an operation process tree in program generation.
For the disjunctive graph, following task decomposition and allocation, we either directly use the LLM to generate the operation execution order or generate dependencies between operations to construct a dependency graph. These two variants are labeled as $(w/ order)$ and $(w/ dependency)$, respectively. Regarding the process tree, we input program generation examples $\mathbb{E}$ directly into the LLM, labeling this variant as $(w/o\ T)$. The quantitative results on 50 tasks with \texttt{GPT-4o} are presented in Tab. \ref{table:ablation}.
The absence of disjunctive graphs leads to a notable decline in \textit{SE}, particularly in multi-robot tasks, indicating that these two variants, which rely entirely on LLM for scheduling, cannot produce feasible and efficient plans for complex tasks due to intricate dependencies. 
Additionally, without a process tree, \textit{Exe} and \textit{GCR} significantly decrease, demonstrating that our tree-based prompts can provide clear guidance, thereby effectively enhancing the executability of programs.

\subsection{Real-world Experiments}
\begin{figure}[!h]
  \centering
  \vspace{0.2cm}
  \includegraphics[width=0.48\textwidth]{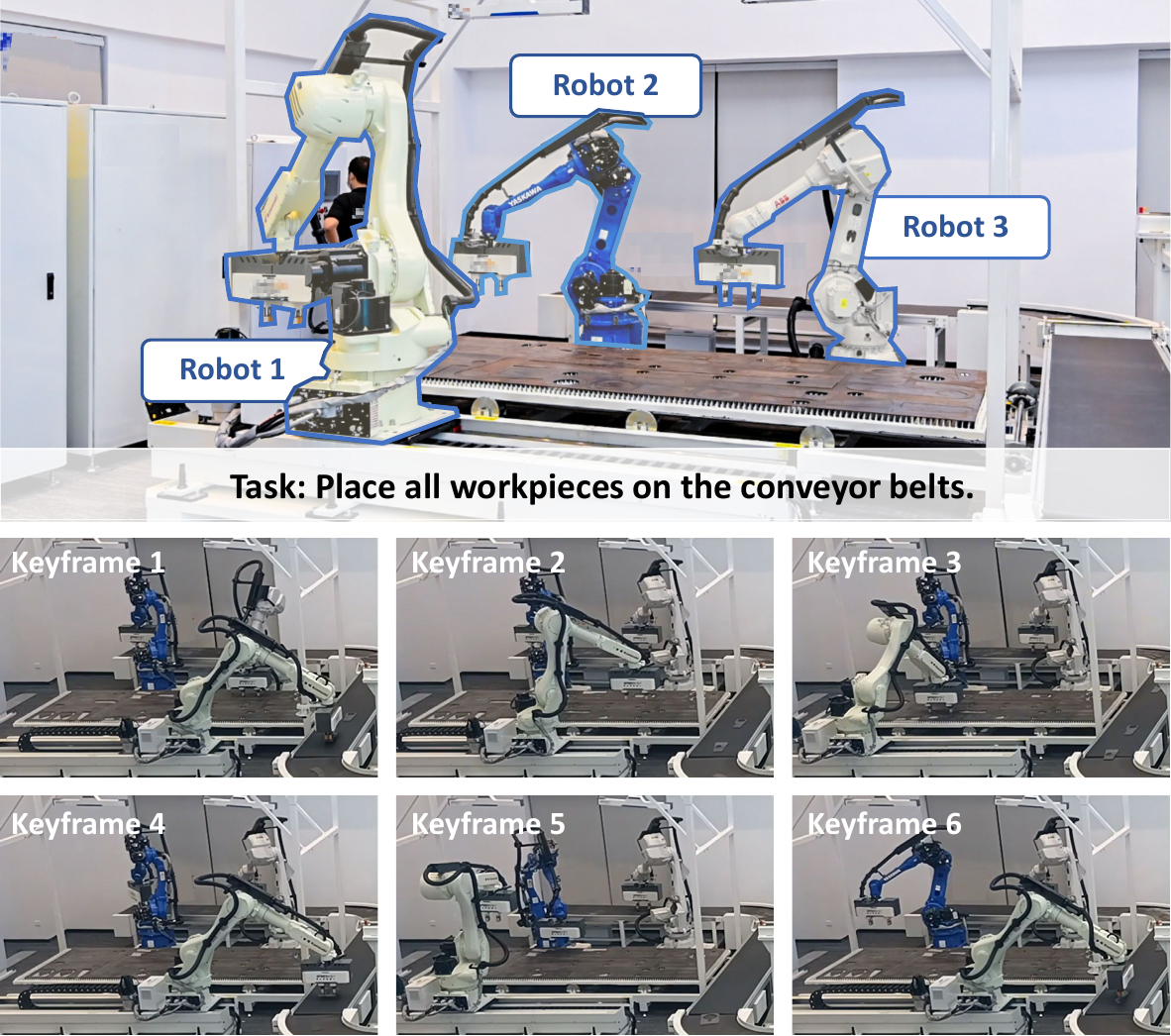}
  \caption{\textbf{A real-world experiment.} Three robots collaborate to complete a transportation task.}
  \vspace{-0.6cm}
  \label{fig:real}
\end{figure}

We also conducted an experiment in a real world scene. 
This scene consists of three robotic arms, each equipped with an end effector for completing transport operations and capable of controlling cameras mounted on a bracket to locate workpieces. 
The three robots collaborate to place workpieces on conveyor belts. To run our method, we used a manually constructed scene description, which took an expert approximately 15 minutes, and then input the scene file and task description to obtain task planning and execution programs. 
After verifying the safety and correctness of the generated results in the simulated environment, we deployed the planning and execution directly into the real scene. The upper part of Fig. \ref{fig:real} presents the setup of the scene, while the lower part displays the key frames during execution. 
The results indicate that our method produces reasonable, efficient plans and correctly executable programs. 

\section{CONCLUSION}

We propose IMR-LLM, a framework for industrial multi-robot task planning and program generation. 
By integrating LLMs with heuristic solvers for task planning and employing operation process trees for program generation, we ensure both efficiency and high executability. 
Experiments on our novel IMR-Bench validate the method's effectiveness in simulated and real environments. 
Future work will incorporate execution feedback \cite{liu2024coherent, yuan2025remac, wang2024llmˆ, huang2022inner, duan2024manipulate} to establish a closed-loop system for enhanced robustness.

\section*{Acknowledgments}

This work is supported by the National Natural Science Foundation of China (62322207, 62325211, 62132021), the Major Program of Xiangjiang Laboratory (23XJ01009), the Key R\&D Program of Wuhan (2024060702030143), and the Natural Sciences and Engineering Research Council of Canada (NSERC).

\clearpage
\bibliographystyle{IEEEtran}
\bibliography{ref}
\addtolength{\textheight}{-12cm}   









\end{document}